\documentclass[letterpaper, 10 pt, conference]{ieeeconf}  

\IEEEoverridecommandlockouts                              

\usepackage{graphicx} 
\usepackage{amsmath} 
\usepackage{amssymb}  
\usepackage[font=small]{caption}
\usepackage{subcaption}
\usepackage{xcolor}
\usepackage{comment}
\usepackage{hyperref}
\let\labelindent\relax
\usepackage{enumitem}
\DeclareMathOperator*{\argmax}{arg\,max}
\title{\LARGE \bf
Visual Prediction of Priors for Articulated Object Interaction
}

\author{Caris Moses*, Michael Noseworthy*, Leslie Pack Kaelbling, Tom\'as Lozano-P\'erez, and Nicholas Roy
\thanks{*Equal contribution.}
\thanks{Computer Science and Artificial Intelligence Laboratory, Massachusetts Institute of Technology, Cambridge, MA, USA}
\thanks{\{carism, mnosew, lpk, tlp, nickroy\}@mit.edu}%
\thanks{We gratefully acknowledge the funding support of the Honda Research Institute and the Education Office at the Charles Stark Draper
Laboratory.
}
}%

\begin{document}

\maketitle
\global\csname @topnum\endcsname 0
\global\csname @botnum\endcsname 0
\thispagestyle{empty}
\pagestyle{empty}

\begin{abstract}

Exploration in novel settings can be challenging without prior experience in similar domains. 
However, humans are able to build on prior experience quickly and efficiently.
Children exhibit this behavior when playing with toys. 
For example, given a toy with a yellow and blue door, a child will explore with no clear objective, but once they have discovered how to open the yellow door, they will most likely be able to open the blue door much faster. 
Adults also exhibit this behaviour when entering new spaces such as kitchens. 
We develop a method, \emph{Contextual Prior Prediction}, which provides a means of transferring knowledge between interactions in similar domains through vision. We develop agents that exhibit exploratory behavior with increasing efficiency, by learning visual features that are shared across environments, and how they correlate to actions. 
Our problem is formulated as a Contextual Multi-Armed Bandit where the contexts are images, and the robot has access to a parameterized action space.
Given a novel object, the objective is to maximize reward with few interactions. 
A domain which strongly exhibits correlations between visual features and motion is kinemetically constrained mechanisms. 
We evaluate our method on simulated prismatic and revolute joints.\footnote{Videos and code are available at \url{https://sites.google.com/view/contextual-prior-prediction}.}
\end{abstract}

\section{Introduction}
\label{sec:intro}
\begin{figure}
     \centering
     \begin{subfigure}{0.23\textwidth}
         \centering
         \includegraphics[clip, trim={0 0cm 0 0cm}, width=\textwidth]{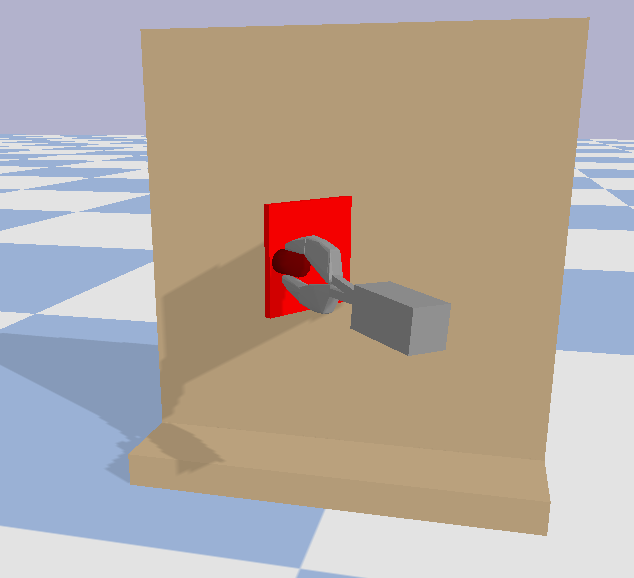}
         \label{fig:slider-gripper}
     \end{subfigure}
     \begin{subfigure}{0.23\textwidth}
         \centering
         \includegraphics[clip, trim={0 3cm 0 5cm}, width=\textwidth]{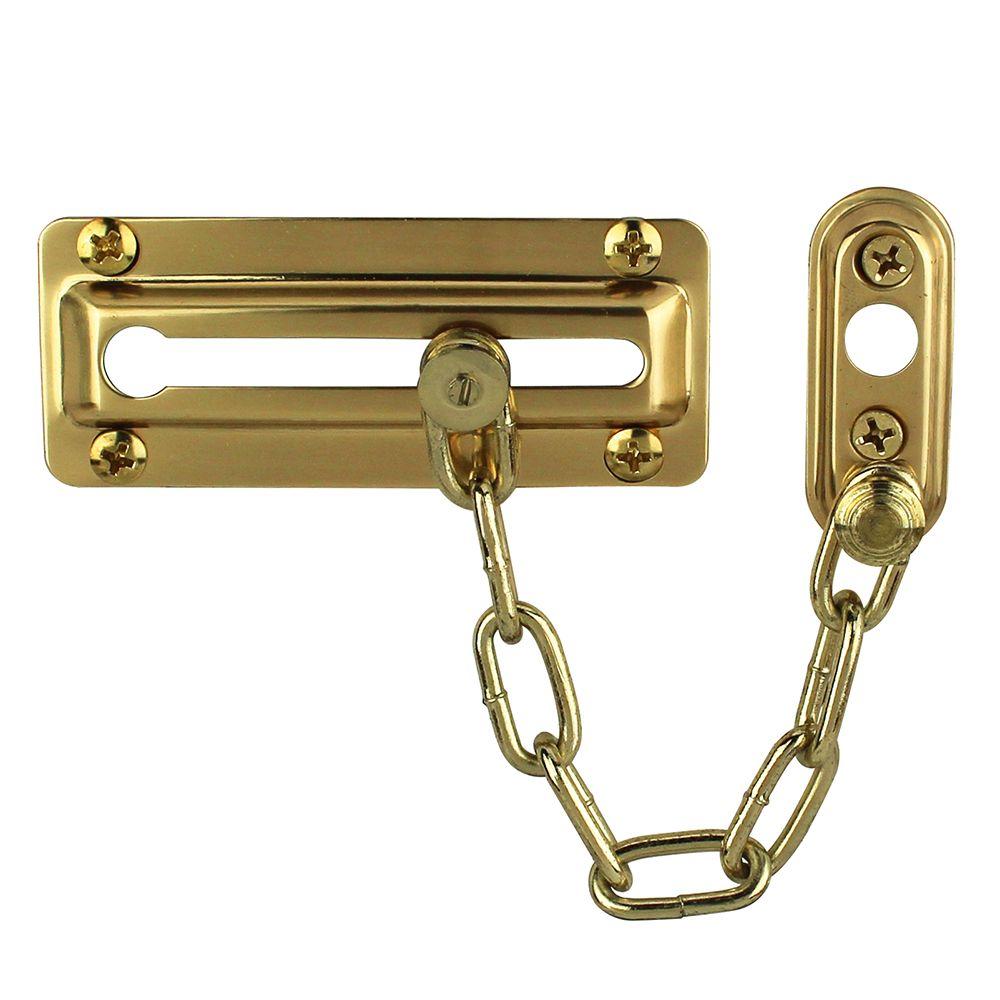}
         \label{fig:door}
     \end{subfigure}
     \vfill
     \begin{subfigure}{0.23\textwidth}
         \centering
         \includegraphics[clip, trim={0 2cm 0 2cm}, width=\textwidth]{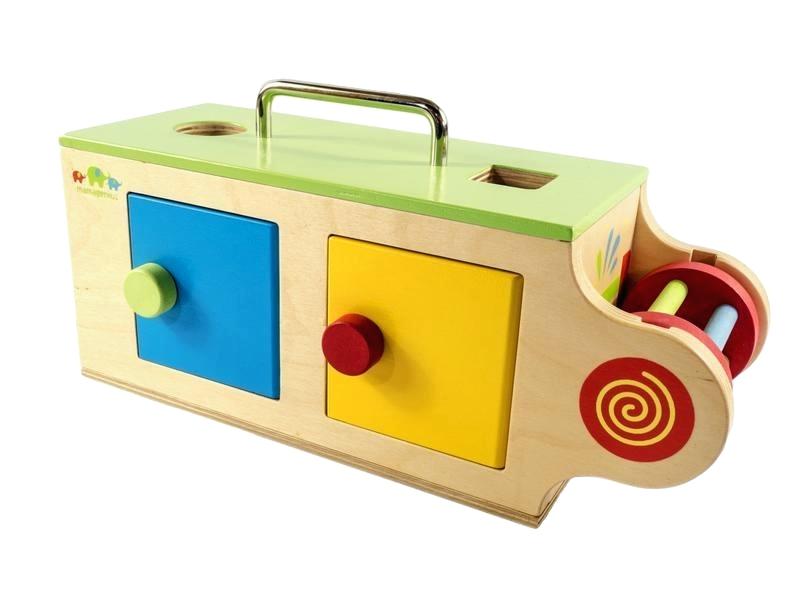}
         \label{fig:toy}
     \end{subfigure}
     \begin{subfigure}{0.23\textwidth}
         \centering
         \includegraphics[clip, trim={4cm 0 4cm 0}, width=\textwidth]{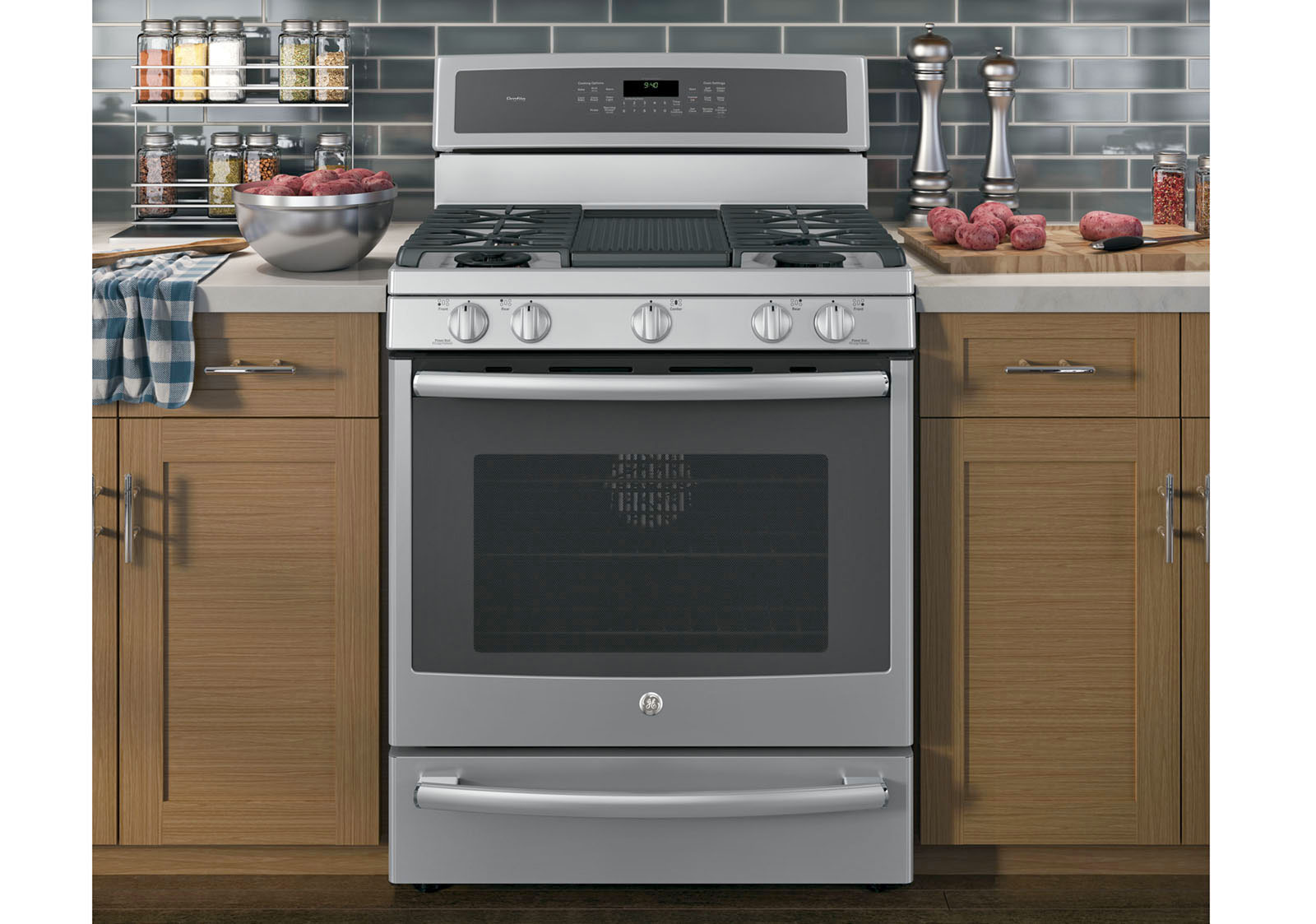}
         \label{fig:oven}
     \end{subfigure}
        \caption{\emph{Contextual Prior Prediction} is motivated by the fact that most objects have rich visual features which indicate their motion. In our simulated door domain (top left), the position of the handle and width of the door indicate the direction which it opens and with what radius. Other objects, such as latches (top right), toys (bottom left), or ovens (bottom right) also have visible indicators such as tracks, hinges, and knobs.}
        \label{fig:example-mechs}
\end{figure}


Humans frequently encounter new objects and are able to successfully articulate them with little to no experience on those particular object instances.
Consider a human entering a new kitchen. 
They can quickly open the cupboards, turn on the lights, and operate the stove, even though they have never used these specific objects before.
This behaviour is enabled by rich visual cues such as the presence of a handle, or the location of hinges on a door (see Figure \ref{fig:example-mechs}). 
These features are useful for inferring the function of a new mechanism and for estimating the motion that it can undergo. 

Our goal is to enable a robot to efficiently interact with novel articulated objects without human guidance by learning from previous visuo-motor experience with related objects.
Previous work has shown how robots can infer the kinematic models of new mechanisms given a single demonstration of the mechanism being actuated \cite{sturm2011probabilistic}.
However, demonstrations are often expensive as they require a human teacher every time the robot needs to interact with a new object.
Other methods provide exploration strategies which enable a robot to estimate the kinematic properties of mechanisms \cite{barragan2014interactive}, \cite{otte2014entropy}.
While these methods perform well, they do not transfer any experience from similar mechanisms when interacting with novel mechanisms.
We desire a solution where the robot uses past experience to experiment efficiently with new mechanism instances to learn how to actuate them from a very small number of self-selected actuations.


Due to constraints present in articulated objects, very few of the possible motion commands the robot can generate  are likely to cause the mechanism to move.
Without models of the object, the robot must propose its own goals or sequences of actions that can quickly generate motion that exhibits the correct kinematic structure of the mechanism.
In order to enable efficient exploration, we propose our method, \emph{Contextual Prior Prediction} (CPP), which uses the visual appearance of a mechanism to provide a prior that indicates which actions are likely to be successful in actuating the mechanism.  The mapping from visual appearance to an actuation prior is learned from previous interactions with mechanisms from the same class.
This enables the robot to only try actions that it believes are likely to succeed based on the new object's appearance.




\begin{figure*}
    \centering
    \small
    \vspace*{4pt}
    \includegraphics[trim={0 0 0 0},clip,width=1\textwidth]{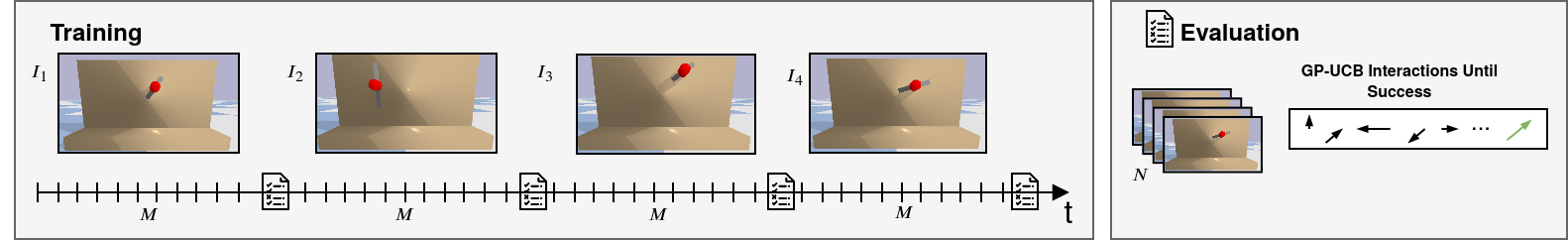}
    \caption{In the learning setup, the robot is sequentially presented with $L$ mechanisms. The robot can interact with each mechanism for $M$ steps (timeline ticks) before a new one appears. After each training mechanism, we evaluate the robot's performance on a separate evaluation set of $N$ novel mechanisms (right). For each evaluation mechanism, the robot takes actions until it has generated an optimal interaction (generated the most possible motion).}
    \label{fig:learning}
\end{figure*}

Thus we have two learning problems.  The {\em outer} problem is to learn a general mapping from the appearance of a mechanism and a proposed action to a {\em reward} which measures how far the mechanism moves, i.e. a reward function. The {\em inner} problem is, given a particular object, and starting with a prediction based on its visual appearance, to efficiently and actively select a sequence of robot motion commands that will cause the mechanism to move.  The system operates in a loop:  given a new mechanism, the robot interacts with it and discovers how it moves; then, the visual appearance of that object together with each of the attempted actions and their effects are added to a training set for the {\em outer} learning problem.  As the robot gets more experience with different mechanisms, it is able to ``understand'' a new mechanism with fewer and fewer trials.

We formulate the overall problem as a \emph{Contextual Multi-Armed Bandit} (C-MAB) in which the robot continuously interacts with a sequence of mechanisms (contexts) with shared structure.  For the {\em outer} problem, we represent the overall reward function using a neural network, which maps visual appearances and possible actions to value, and train it using conventional supervised-learning methods.  Then, for the {\em inner} problem, given a novel mechanism to interact with, we use the neural network to predict the expected value of each possible action in a continuous action space.   We treat this function as the prior mean of a \emph{Gaussian Process} (GP), and use the \emph{Gaussian Process Upper Confidence Bound} (GP-UCB) strategy \cite{gpucb} to handle the exploration-exploitation tradeoff when finding the optimal action. 

In our method, GP-UCB is also used in the data collection phase. In this paper we explore the effectiveness of GP-UCB and random sampling as exploration strategies during training. At evaluation time, we compare the overall effectiveness of the system to one that applies GP-UCB to each new mechanism starting from a generic prior, as well as to a baseline random search.  The techniques are generic and could apply to a variety of mechanisms.  Our experimental comparisons are done in a simulated domain containing prismatic and revolute joints with different visual appearances and kinematic parameters.

The contribution of this paper is to demonstrate how a period of exploration with mechanisms with different visual appearances can lead to the ability to actuate never-before-seen mechanism instances with very few (sometimes just one) trials.  In addition, the approach of learning to map appearance to the prior of a GP, rather than mapping appearance directly to motor commands, means that the overall system is significantly more robust, and can recover from inaccurate predictions that arise when there is little training data.


\section{Method}
\label{sec:method}

In this section we introduce \emph{Contextual Prior Prediction} and discuss how CPP can be used to find optimal actions in novel contexts. 
Section \ref{sec:method-problem} describes how CPP fits into the C-MAB framework. 
Section \ref{sec:method-gpucb} reviews GP-UCB, a method which can be used to learn a context-specific reward function, and how we extend this method to C-MABs by learning a prior on the context-specific reward function.
In section \ref{sec:method-reward} we discuss the structure of our prior, and how it enables visual feature learning.

\subsection{Problem Formulation}
\label{sec:method-problem}

Our problem can be formulated as a \emph{Contextual Multi-Armed Bandit} in which a robot is given a context, $I$, to interact with, then selects an action with the objective of maximizing its reward. We assume the actions are a continuously parameterized set of motion primitives, e.g., a pushing or twisting motion, where the parameters encode variations such as the direction of force, contact point, axis of rotation, etc. As a result, the action space is typically a continuous-valued domain such as $\mathbb{R}^d$ or SE($d$). 
Our setup differs from the typical C-MAB in that the robot will get several interactions with each context.
CPP provides a way to leverage previous experience, in the form of a prior on the context-specific reward function, when interacting with novel contexts. 
We evaluate our method in a
learning framework in which there are separate training and evaluation phases, as shown in Figure~\ref{fig:learning}.

During the training phase, the robot is sequentially given $L$ different mechanisms and is allowed to select $M$ interactions with each, and observe the rewards, that is, the degree to which the robot was able to actuate the mechanism.  This data is used to train a model.  We then carry out evaluation trials on $N$ novel contexts:  in each one, the robot is able to interact until it fully actuates the mechanism. In each evaluation context, $I_n$, the robot uses the learned model as a prior on the context-specific reward function to quickly maximize it.
To measure the robot's success we assume that an oracle is able to provide the optimal reward, $r^*_n$, and we calculate the {\em normalized simple regret}, $e$, which is the loss of not selecting the optimal action. Here, $r$ is the reward for the robot's chosen action.

\begin{equation}
    e = \frac{r^*_n - r}{r^*_n}\;\;
    \label{eq:regret}
\end{equation}

Therefore, during each evaluation, we count the number of interactions needed until the robot can achieve regret less than a specified threshold ($e < 0.05$ in our experiments).


More formally, let ${\cal I}$ be the space of possible contexts and ${\cal A}$ be the space of actions that the robot can execute.  We are interested in learning reward function $R: {\cal I} \times {\cal A} \rightarrow \mathbb{R}$, which specifies the value of taking an action given an image.  We will learn an approximation of $R$, represented as a \emph{Neural Network} (NN), denoted $f_\text{NN}$. 

\subsection{Gaussian Process Optimization with a Learned Prior}
\label{sec:method-gpucb}

At evaluation time, if we had high confidence in the NN's prediction of reward, we would, given a new image context $I$, simply execute the action

\begin{equation}
a^* = \argmax_{a \in {\cal A}} f_\text{NN}(I, a)\;\;.    
\end{equation}

However, when we have a small amount of experience that prediction may not be very accurate, particularly in novel contexts. We would like to extend the model such that when the robot takes actions and fails, the reward from failed attempts inform future interactions as the robot searches for the optimal action.

The \emph{Gaussian Process} provides one method for inferring the value of some unknown function $f \sim GP(\mu, k)$, 
where $\mu$ is the mean function defined on domain ${\cal A}$, $f$ is a function on that same domain, and $k$ is a kernel function which models the covariance between pairs of function values depending on the domain points at which the function is evaluated~\cite{rasmussen2003gaussian}. Given $t-1$ samples of action and reward pairs $(a_{0:t-1}, r_{0:t-1})$, the GP provides a Gaussian posterior $\mathcal{N}(\mu_{t-1}(a), \sigma_{t-1}(a))$ over the reward of the next action $a_t$ for any action $a_t \in {\cal A}$. 

A simple procedure would be to take the action with the largest mean reward given the actions so far, but GP-UCB combines the mean and the variance to allow the robot to choose actions that have the potential for high reward due to a high degree of uncertainty in the model. The GP-UCB criterion \cite{gpucb} generates an action sample $a_t$, given both the prior and the values of the previous samples, according to the objective
\begin{equation}
    a_{t} = \argmax_{a \in {\cal A}} \mu_{t-1}(a) + \beta^{1/2}_t \sigma_{t-1}(a)
\end{equation}
where $\mu_{t-1}(a)$ and $\sigma_{t-1}(a)$ are the mean and variance of the function value at $a$ at time $t-1$, and $\beta$ is a parameter which trades off between sampling parameters with a high mean and those with high variance. 

Typically the mean function is initialized to zero, $\mu_0(a) = 0$, but we have the pre-trained NN to provide guidance before any new data is collected. 
To incorporate the NN, we use the GP to model the residual function $f_\text{RI}$ between the true reward function and our prediction for context $I$, so $f_\text{RI}(a) = R(I, a) - f_\text{NN}(I,a)$.   
On iteration $t$ of the GP-UCB optimization process, we select

\begin{equation}
\label{eq:gpucb}
    a_t = \argmax_{a \in {\cal A}}  f_\text{NN}(I, a) + \mu_{t-1}(a) + \beta^{1/2}_t \sigma_{t-1}(a).
\end{equation}
The GP will have a prior mean function that is constant 0, and a fixed kernel function, but the above function is equivalent to putting a GP on $R(I, \cdot)$ with prior mean function $f_\text{NN}(I,\cdot)$.
Under this criteria, samples initially come from parts of the space where the learned reward function predicts high reward. Then as the true context-specific reward function is learned, we select actions with much more accurate knowledge of the true underlying reward function.

In this framework, to select actions according to the GP-UCB criteria given a context $I$, we must optimize Equation \ref{eq:gpucb}.
To do this we start by evaluating the function on a coarse sampling of the action space.
We then perform non-linear optimization on a few of the best samples, and select the best optimization run as the criteria-maximizing action.
To find the agent's current best estimate of the optimal action, in order to calculate Equation \ref{eq:regret}, we follow the same procedure but with $\beta=0$.

In our experimental results we use a \emph{squared exponential} kernel and tune the kernel parameters by executing GP-UCB on a separate set of random mechanisms and observing the resulting exploration strategies.
We aim to find a good balance between exploring areas of the input space with high uncertainty and areas with known high reward.

\subsection{Learned Reward Function}
\label{sec:method-reward}

We approximate $f_\text{NN}$ with a NN which consists of independent encoders for the input channels (images and action parameters) and a regressor which uses these encodings to predict reward.
The image encoder $f_\text{im}$ has the form $z_\text{im} = f_\text{im}(I) = f_\text{ss}(f_\text{cnn}(I))$, where $f_\text{cnn}$ is a \emph{Convolutional Neural Network} (CNN).
We found that mapping from the CNN directly into a fully connected layer did not result in useful encodings, so we added $f_\text{ss}$, which is a {\em spatial softmax} layer~\cite{finn2015learning}.    It generates, as output, a set of 2D feature points that are salient for making value predictions.
Each 2D feature point is the expected pixel location of activations in one of the final CNN channels.

For the action inputs, a \emph{Multi-Layer Perceptron} (MLP), $f_\text{a}$, transforms the action parameters into a latent space.   This part of the network was designed so that additional action types could be added to the system, in which case, action parameters for all policies would be transformed into a shared space.  This yields action encoding 
    $z_a = f_\text{a}(a)$.
Finally, 
$z_{im}$ and $z_a$ are  concatenated and passed through an MLP, $f_\text{dist}$, which learns to predict how the action and image features map to a reward.
The NN is composed of these encoders and the distance regressor as follows,
\begin{equation}
    f_\text{NN}(I, a) = f_\text{dist}([f_\text{im}(I); f_\text{a}(a)])\;\;.
\end{equation}

If the NN is trained to effectively approximate the true reward function $R$ then an approximately optimal policy has the form
\begin{equation}
    \pi(I) = \argmax_{a \in {\cal A}} f_\text{NN}(I, a)\;\;.
\end{equation}
This formulation suffices for mechanisms that can be effectively actuated by a single relatively simple parameterized action. 
If we were to move to truly sequential mechanisms, such as gate latches, it would be necessary to treat the problem as a \emph{Contextual Markov Decision Process} rather than a contextual bandit.

\section{Experiments}
\label{sec:exp}
\newcommand{\ourmethod}{CPP}
\newcommand{\ourmethodlong}{contextual prior prediction}

In this section we describe the effectiveness of \ourmethod{} in the \emph{Slider} and \emph{Door} domains.
We find that \ourmethod{} is able to learn from relatively few training mechanisms to quickly operate a new mechanism.
We further analyze different data generation methods and their performance.
Our experiments use the \emph{PyBullet} simulator \cite{coumans2019}.

\begin{figure*}[t!]
\begin{subfigure}{0.49\textwidth}
    \includegraphics[trim={0cm 0 0cm 0},width=\textwidth]{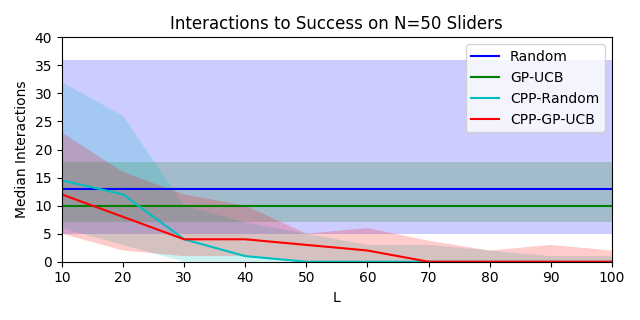}
\end{subfigure}
\begin{subfigure}{0.49\textwidth}
    \includegraphics[trim={0cm 0 0cm 0},width=\textwidth]{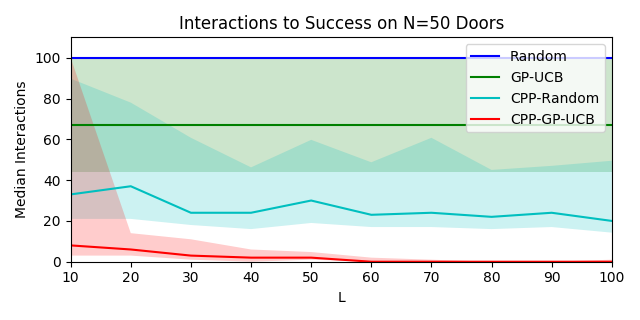}
\end{subfigure}
\caption{Number of interactions until success (less than $0.05$ regret) on $N$ novel sliders (left) and doors (right). The median number of interactions is reported for $N=50$ evaluation mechanisms for $5$ separately trained NN models. The plots show performance for models that have been previously trained on $L$ mechanisms (x-axes) each with $M=100$. We compare our method, noted as CPP-GP-UCB and CPP-Random, to GP-UCB which does not learn from previous interactions, and a random baseline, Random. 25\% and 75\% quantiles are plotted.}
\label{fig:regret}
\end{figure*}

\begin{figure}[t!]
\includegraphics[trim={0cm 0 0cm 0cm},width=0.48\textwidth]{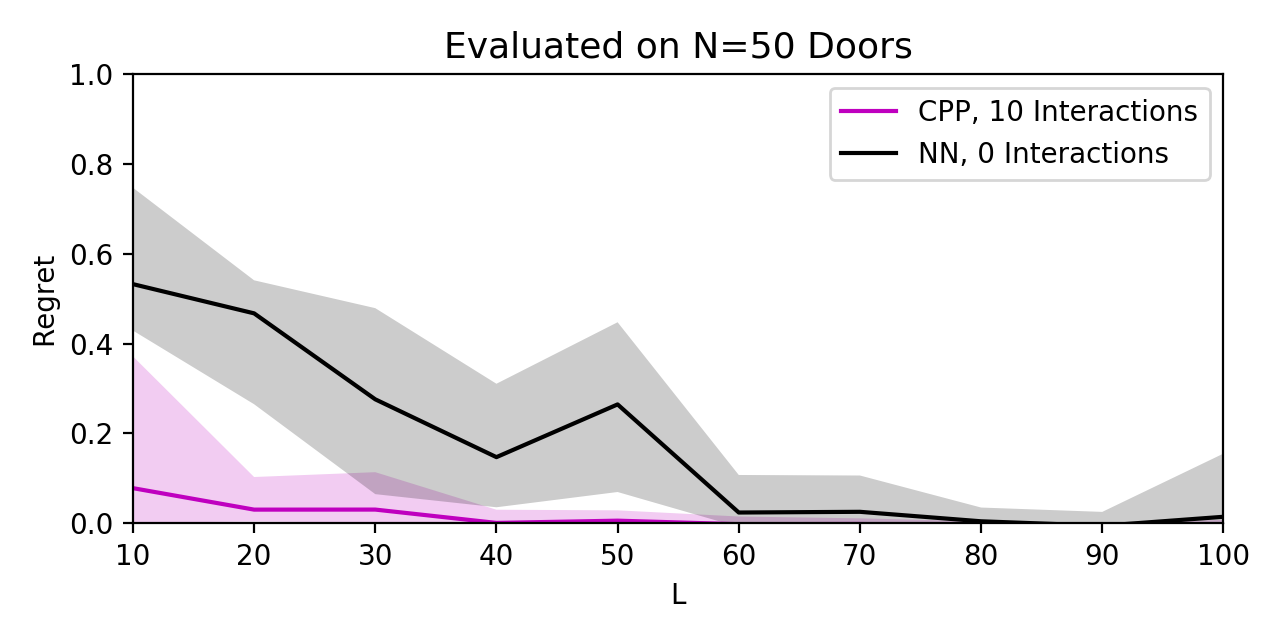}
\caption{This plot compares using the NN to directly predict optimal actions, to using 10 GP-UCB interactions on top of the NN to find optimal actions (\ourmethod{}). As shown, the pure NN initially results in poor performance as compared to \ourmethod{}.}
\label{fig:nn-vs-cpp}
\end{figure}

\subsection{Domains}
\label{sec:exp-bb_domain}

\subsubsection{Sliders}
Sliders are prismatic joints that vary in length, position, and the angle of the slider's track.
Their action space, $\mathcal{A}_{prism}$, consists of parameterized actions that actuate prismatic joints.
Prismatic joints are parameterized by the pose of the origin, $\boldsymbol{a} \in SE(3)$, a unit vector direction in the frame of the origin, $\boldsymbol{e} \in \mathbb{R}^3$, and the desired configuration, or distance to move the slider handle, $q \in \mathbb{R}$. 
We search the space of $q$ and the pitch component of $\boldsymbol{e}$.

\subsubsection{Doors}
Doors are revolute joints that vary in size and the direction in which they open.
Their action space, $\mathcal{A}_{rev}$, consists of parameterized actions that actuate revolute joints which are parameterized by the pose of center of rotation, $\boldsymbol{c} \in SE(3)$, the distance from the center of rotation along the $x$-axis to the mechanism's handle, and $r \in \mathbb{R}$, and the desired configuration, or opening angle of the door, $q \in \mathbb{R}$.
We search the space of $r$, $q$, and the pitch component of $\boldsymbol{c}$.

With \emph{PyBullet}\cite{coumans2019} we can generate multiple mechanisms where instances of the same joint type share visual structure.
In our experiments, the robot will see a sequence of randomly generated mechanisms belonging to the same class.
See Figure \ref{fig:learning} for example sliders and Figure \ref{fig:example-mechs} for a door. 

An action outputs a trajectory the end-effector should follow to actuate a joint with the corresponding parameters.
Given the pose of the mechanism handle, and action parameters, a trajectory is generated in the mechanism's configuration space.  
Then inverse-kinematics and Cartesian interpolation are used to generate a trajectory in Cartesian space.
This trajectory is executed using a PD controller by applying forces to the handle. Due to the mechanism's constraints, the applied forces do not always result in motion.

Training and evaluation are interleaved. As the robot interacts with \emph{training} mechanisms, it is intermittently evaluated on a new set of \emph{evaluation} mechanisms.  It gets to interact with each new evaluation mechanism until it is able to maximize its reward.
In all our experiments, the reward is the distance the mechanism moves.

\subsection{Interacting with a New Mechanism}
\label{sec:exp-overall}

In this section we compare methods for interacting with a new mechanism. The \ourmethod{} method can work with any kind of exploration strategy during training time. 
In Figure~\ref{fig:regret} we show our method using two different exploration strategies during training time. 
We will discuss the comparison of these two methods more in Section \ref{sec:exp-data}.

\begin{description}
\item{\textbf{CPP-Random}}: We randomly sample from the action space to collect training interactions for each mechanism.

\item{\textbf{CPP-GP-UCB}}: We use the GP-UCB algorithm (see Section \ref{sec:method-gpucb}) to collect training interaction data. With this method the agent is actively trying to maximize its reward with each mechanism.

\end{description}

We compare our method against two simple but sensible baseline methods for evaluation that do not try to use previous experience and visual information about the new mechanism to predict how to actuate it.  
Thus, for the baseline methods, each mechanism is a new problem. While we expect performance to improve {\em within} one trial of interaction with a mechanism, we do not expect performance to improve {\em across} interactions with different mechanisms.

\begin{description}
\item{\textbf{Random}}: We randomly sample from the action space until the agent is able to maximize its reward.

\item{\textbf{GP-UCB}}: The robot uses the GP-UCB algorithm (see Section \ref{sec:method-gpucb}) for action selection until it is able to maximize its reward.
\end{description}

In both the baselines and in our evaluation of \ourmethod{}, we limit the agent to 100 attempts at maximizing its reward.
The results of our experiments are shown in Figure~\ref{fig:regret}.
Each plot shows the number of steps each method took to maximize the reward on the evaluation mechanisms as a function of $L$, the number of mechanisms the robot interacted with during training time. 
The baseline methods (blue and green) have the same median for all $L$ values because they are not leveraging previous experience, and thus cannot show improvement.  

The learning-based methods (red and cyan) show significant decreases in the number of interactions required to generate a successful interaction. 
Not only does the median number of interactions to success decrease with $L$, but so do the quantiles, meaning these methods more reliably interact with novel mechanisms.
The larger $L$ values are important to us:  a well-trained robot is able to actuate a mechianism without any experimentation!

One drawback of \ourmethod{} is that it can be misled by a poorly trained NN.
In this case, the robot will first explore areas where the NN predicts high reward even if it is wrong. 
The GP-UCB algorithm will eventually correct the model's beliefs and explore other regions but this may take longer than an uninformed prior.
We note that in all cases, as the NN starts performing better (after seeing more unique mechanisms), the number of required interactions decreases.

To visualize the usefulness of CPP versus just trusting our NN predictions, we compare an agent that simply selects the best action according to the NN, to one that that uses our \ourmethod{} method. 
Figure \ref{fig:nn-vs-cpp} shows a \ourmethod{} agent which performs 10 GP-UCB interactions on top of the learned visual prior. 
As shown, \ourmethod{} can still achieve low regret even with a poor NN prior.

\begin{figure*}
    \centering
    \vspace{2mm}
    \includegraphics[width=0.95\linewidth]{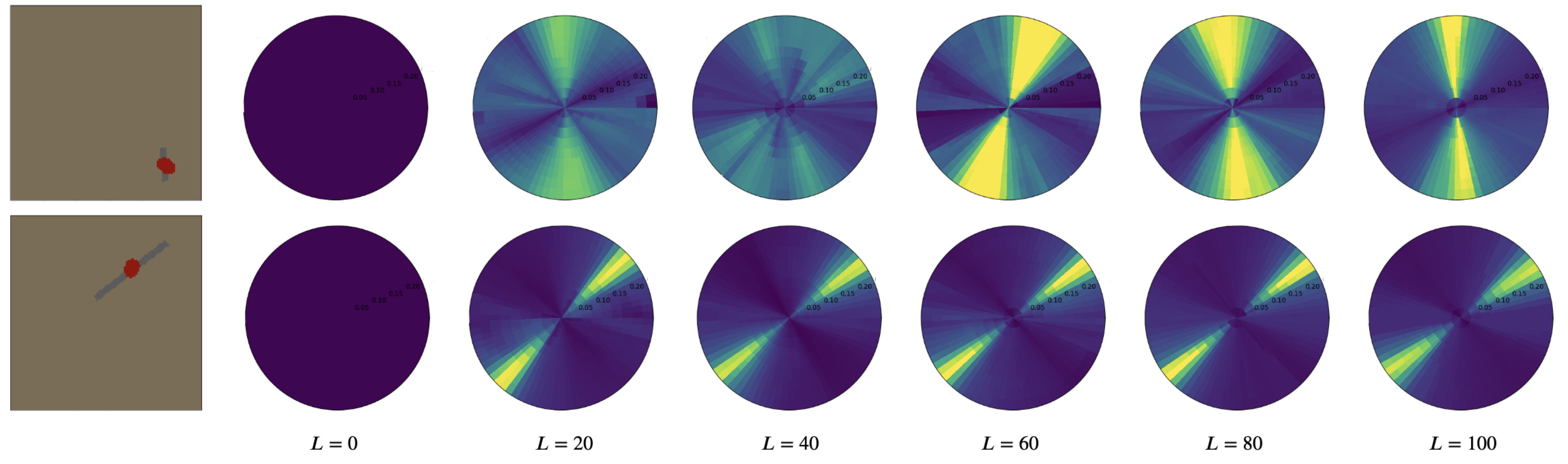}
    \caption{Visualization of the NN prior after experience with $L$ previous sliders for $2$ novel sliders. Each plot visualizes the predicted reward for an action, given in polar coordinates (direction and distance to move the handle). Yellow indicates a higher predicted distance. In all rows the predictions improve as $L$ increases. However they all improve at different rates. This is most likely a factor of how closely these evaluation sliders correspond to sliders seen during training. }
    \label{fig:prior-viz}
\end{figure*}

To visualize how the NN prior improves over time, we show its predictions after being trained on an increasing number of sliders in Figure~\ref{fig:prior-viz}.
The predictions get better at different rates for each slider which likely correlates to how similar the evaluation sliders are to the training sliders.

\subsection{Generation of Training Interactions}
\label{sec:exp-data}

\begin{figure}[t]
\centering
\begin{subfigure}{0.23\textwidth}
    \includegraphics[trim={0cm 0 0cm 0},width=\linewidth]{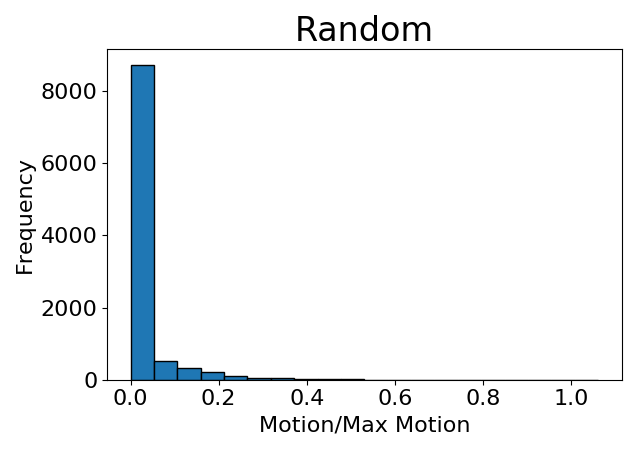}
\end{subfigure}
\begin{subfigure}{0.23\textwidth}
    \includegraphics[trim={0cm 0 0cm 0},width=\linewidth]{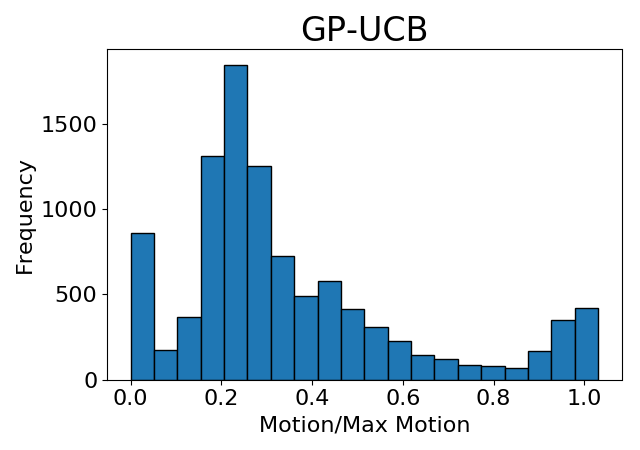}
\end{subfigure}
\caption{Motion histograms of the data collected for $M=100$ steps on $L=100$ doors using random data collection (left) and GP-UCB active data collection (right). The GP-UCB sampling is biased toward collecting samples with more motion.}
 \label{fig:door-hists}
\end{figure}

The primary utility of GP-UCB is to generate useful interactions at evaluation when the learned NN does not have accurate predictions. We also experimented with using GP-UCB for collecting useful actions while interacting with training mechanisms. 


Figure \ref{fig:regret} shows that both the CPP-Random and CPP-GP-UCB methods perform similarly in learning a good prior for slider mechanisms.
However, for door mechanisms, we see that the CPP-GP-UCB method outperforms CPP-Random. 
This is due to the size of the action space, the size of the rewarding region of the action space relative to the size of the entire space, and the complexity of the reward function (how dependent the policy parameters are on each other).
Figure \ref{fig:door-hists} gives a histogram visualization of the training data used for door mechanisms.
The random exploration strategy generates mostly zero motion, while the GP-UCB method is able to effectively actively explore the action space to find rewarding samples useful for training the NN.

\subsection{Baxter Proof of Concept}
\begin{figure}
    \centering
    \includegraphics[clip, trim={0 1.8cm 0 0}, width=0.38\textwidth]{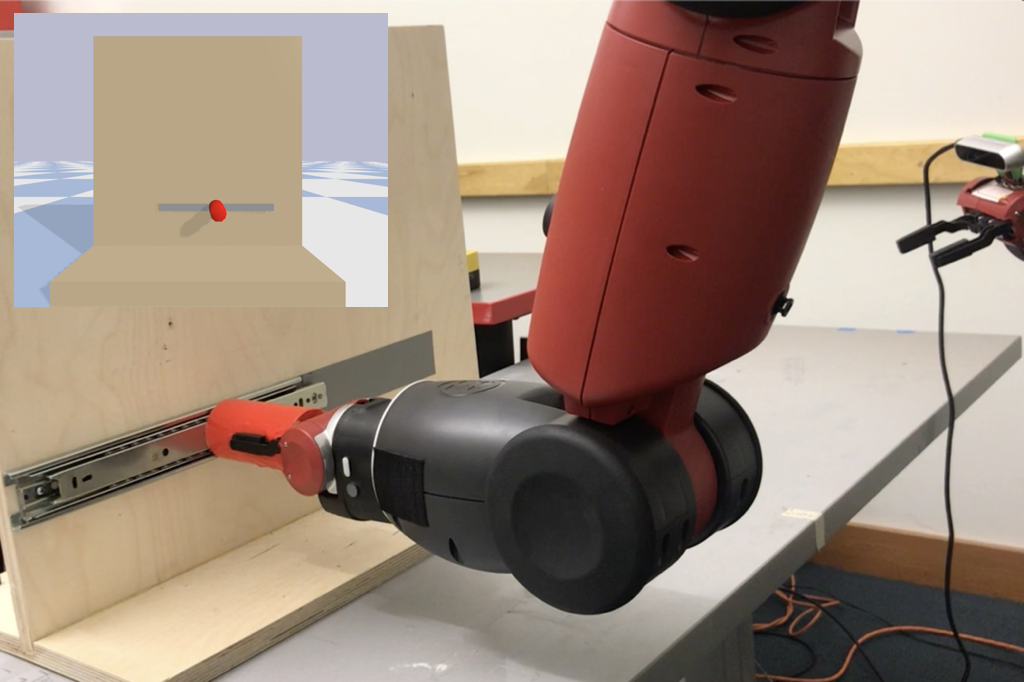}
    \caption{The real  and simulated (inset) slider actuated by the Baxter.}
    \label{fig:realbb}
\end{figure}

We tested our learned model for slider mechanisms on a Baxter robot. The policy parameterizations are the same, and the trajectories output by the policies are fed into a position controller for the Baxter end effector, as seen in Figure \ref{fig:realbb}. 
Our objective was to determine if the evaluation part of our pipeline could be executed on a real robotic platform. 
We observed that the Baxter was able to explore the real, novel slider mechanism when it started from both a poor (little previous experience) and good NN model. 
The input image is a simulated version of the real mechanism as depicted in the inset of Figure \ref{fig:realbb}.
We surprisingly found that the Baxter was able to generate more motion for imperfect actions than the simulated agent.
This was due to the compliance in the Baxter arm, which actually aided in shaping the reward, enabling the Baxter to learn the correct slider parameters with very few interactions.
The robot interactions can be seen in the accompanying video.

\section{Related Work}
\label{sec:related}
Our work lies at the intersection of estimating kinematic models and policy learning. While the former can be used in the latter, there has been little work tying the two together with the goal of generalizing to novel objects through vision. 

A recent area of interest, which our work falls into, is learning  low-level policies from pixel inputs. 
In this setting, ideally a robot could be trained to do any task with basic sensing and the right learning framework, eliminating the need to manually engineer a control system.
Finn et al. \cite{finn2015learning, levine2016end} train a network to perform various manipulation tasks from pixel-space input. 
Their work focuses on learning to perform a single task well, whereas we focus on manipulating multiple objects that share structure.
Like Agrawal et al. \cite{agrawal2016learning}, we use a model-based policy.
Whereas they learn the forward and inverse models for poking objects, we assume we have a structured policy space based on kinematic models. Other model-based work attempts to learn a dynamics model to be used in a controller \cite{lenz2015deepmpc,fragkiadaki2015learning, nagabandi2018neural}.

Recent work has started to explore the idea of learning an embedded space of policies to handle the more general case of policy learning for a variety of tasks in a single environment. In \cite{lynch2019learning} they learn from teleoperated ``play'' data in which a human teleoperates a simulated robot to generate data to train an embdedded policy space. In \cite{hausman2018learning} they have a similar objective, but the data is generated via off-policy reinforcement learning. In both works the agent is constrained to perform well in only a single environment.

Another approach to learning generalizable policies is to explicitly incorporate properties that represent shared structure (e.g., dynamic, kinematic, static \cite{bergstrom2011scene}) into the model. 
Of particular relevance to our work are methods that estimate the kinematic parameters of articulated objects from interaction data \cite{sturm2011probabilistic, barragan2014interactive, otte2014entropy, katz2008manipulating, hausman2015active, eppner2017physics}.
However, these works focus on learning the kinematic parameters for a single object.
In \cite{abbatematteo2019learning}, the authors learn to predict the kinematic parameters of a class of objects from an image.  
These works use visual representations such as optical flow  \cite{bergstrom2011scene, katz2008manipulating}, or point features \cite{eppner2017physics}, which have been shown to be useful in predicting either kinematic models \cite{katz2008manipulating, eppner2017physics} or rigid body separation \cite{bergstrom2011scene}.
Like \cite{abbatematteo2019learning}, we use a CNN to learn visual features that are informative of motion in a class of objects.

A closely related problem is that of learning object affordances which is primarily concerned with the effects of actions as opposed to their rewards. 
In \cite{nguyen2016detecting}, the authors predict a probability distribution over possible object affordances from pixels. In \cite{montesano2009learning}, the authors learn the probability of successfully executing a grasp. Our method extends similar lines of work, in that the agent uses the learned model to seed on-line interactions.

GPs are useful for active exploration due to the fact that they give an estimate of the uncertainty over predictions \cite{gpucb}. 
However, using GPs requires careful specification of the kernel function which can be difficult to specify for images \cite{kapoor2010gaussian, van2017convolutional}. 
In CPP we are able to reap the benefits of using a GP for exploration by using the image to initialize a GP over the low dimensional action space.


\section{Conclusion}
In this work, we focus on the problem of efficiently exploring novel mechanism instances by transferring knowledge from interactions with previous mechanisms through vision.
We develop a method, \emph{Contextual Prior Prediction}, that uses a NN as a prior mean for a GP which is used for exploration.
We evaluate our method in a continual C-MAB learning framework on a simulated domain consisting of prismatic and revolute joints, and prove that the evaluation strategies can be executed on a real robotic platform.
As the robot interacts with more mechanism instances, it can successfully actuate a new mechanism with an increasingly smaller number of interactions.
Future work needs to be done evaluating what would be necessary to learn relevant features for predicting motion from realistic images.
We also would like to extend our method to more complex domains which require sequential manipulation.


\bibliographystyle{IEEEtran}
\bibliography{IEEEabrv,refs}



\end{document}